# Deep Learning for Lip Reading using Audio-Visual Information for Urdu Language

Muhammad Faisal
Information Technology University
Lahore
m.faisal@itu.edu.pk

Sanaullah Manzoor
Information Technology University
Lahore
sanaullah.manzoor@itu.edu.pk

**Abstract**

*Human lip-reading is a challenging task. It requires not only knowledge of underlying language but also visual clues to predict spoken words. Experts need certain level of experience and understanding of visual expressions learning to decode spoken words. Now-a-days, with the help of deep learning it is possible to translate lip sequences into meaningful words. The speech recognition in the noisy environments can be increased with the visual information [1]. To demonstrate this, in this project, we have tried to train two different deep-learning models for lip-reading: first one for video sequences using spatio-temporal convolution neural network, Bi-gated recurrent neural network and Connectionist Temporal Classification Loss, and second for audio that inputs the MFCC features to a layer of LSTM cells and output the sequence. We have also collected a small audio-visual dataset to train and test our model. Our target is to integrate our both models to improve the speech recognition in the noisy environment.*

## 1. Introduction

The ability of recognizing what is being said only from visual information is an impressive skill but a difficult task for the novice. The task of lipreading is inherently ambiguous at the word level and short sentences, because of the absence of the context and secondly some characters that produce the same lip sequences, called homophemes, are difficult to distinguish. But this difficulty can be overcome using the context of neighborhood information. The neighboring words can help to minimize the ambiguity of homophemes.

Performance of automatic speech recognition system (ASR) can be increased with the help of visual information in the noisy environments.

But human lipreading performance is not precise consequently there is enormous need of automatic lip-reading system. It has many practical applications such as dictating instructions or messages to a phone in a noisy environment, transcribing and re-dubbing archival silent films, security, biometric identification, resolving multi-talker simultaneous speech, and improving the performance of automated speech recognition in general [1, 2].

Usually in noisy environments speech recognition systems fails or performs poorly, because of the extra noise signals. To overcome this problem and improve the performance of speech recognition system in noisy environments we can add the visual information. The visual information (lips movements of the speaker) can help to speech recognition systems. There are many challenges that make the lipreading task very difficult some of the major challenges are:

- Absence of context
- Extraction of spatio-temporal features
- Some people are more expressive with their lips while others are not (visually-speechless-person)
- Generalization across Speakers
- Guttural sounds (like consonants K and G)
- Mumbling sounds

In this project, we trained two separate deep learning based models, the first model is LipNet [2], which was trained on GRID dataset we retrained it on our own dataset of Urdu language. This model is end-to-end sentence level lipreading model. This model operates at the character level using the spatio-temporal convolutions (STCNNs) recurrent neural networks (RNNs) and finally the connectionist temporal classification loss (CTC) [3]. The second network processes the audios features using the LSTMs and categorical cross entropy loss.

The next section summarizes the literature review. In section 3 we briefly described the network architecture and our dataset is explained in section 4. The implementation details are explained in section 5. Finally the experiments and results are discussed in section 6.

## 2. Literature Review

In this section, we outline several approaches to automatic lipreading and audio-visual lipreading.

### 2.1 Lip Reading

A large body of work has been done on lip reading using pre-deep learning methods. Many approaches using the convolutional neural networks have been proposed to recognize phonemes [4] and visemes [5] from still images of lips movement, instead of recognizing full words and sentences. A phoneme is the smallest distinguishable unit of sound that collectively make up a spoken word; a viseme is its visual equivalent (lips movement).

Recently a deep learning based lip-reading model has been proposed called LipNet [2], that consists of three spatio-temporal convolutional layers, followed by bi-directional gated recurrent units (Bi-GRU), and finally a connectionist temporal classification (CTC) loss. They reported 96 % accuracy on the GRID dataset.

### 2.2 Audio-visual Speech Recognition

The audio-visual speech recognition system is very similar to the lip-reading and their problems are closely linked. [6] used feed-forward Deep Neural Network (DNNs) to perform phoneme classification using a large non-public audio-visual dataset. Recently, [1] has proposed a large and complex end-to-end trainable network for audio-visual speech recognition; their model consists of two encoders and one decoder. They named their model as Watch, Listen, Attend and Spell (WLAS) network. The first encoder encodes the video frames by passing them through convolutional layers followed by stacked LSTMs. A fixed size encoded vector is stored. The second encoder encodes the audio MFCC features by passing it through the stacked LSTMs. Later on, these both encoded vectors are concatenated and input to decoder networks that also consists of stacked LSTMs, the decoder network outputs the character sequence.

## 3. Network Architecture

Lip-reading architecture has two networks, one network is used for lip-reading context prediction and second network is deployed for speech recognition. In the following section, details of both networks are reported. Figure 1 shows the lip-reading network and Figure 2 illustrates the configuration of our speech recognition network.

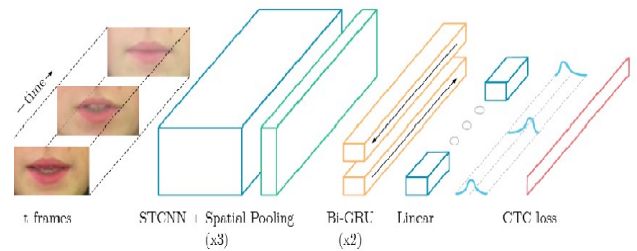

Figure1: LipNet architecture. A sequence of T frames is used as input, and is processed by 3 layers of STCNN, each followed by a spatial max-pooling layer. The features extracted are processed by 2 Bi-GRUs; each time-step of the GRU output is processed by a linear layer and a softmax. This end-to-end model is trained with CTC [2].

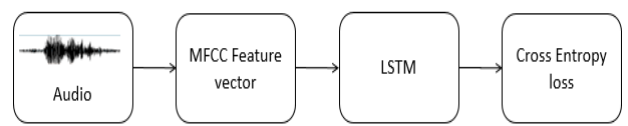

Figure 2: Architecture of Speech Recognition Network

### 3.1 Spatiotemporal Convolution based Lip-Reading Context

Convolutional neural networks (CNNs) with stacked convolutional layers are instrumental in performance in computer vision based tasks such object recognition [10]. A basic 2-Dimensional convolution layer expression is given by

$$[conv(x,w)]_{c'ij} = \sum_{c=1}^{C} \sum_{i'=1}^{k_w} \sum_{j'=1}^{k_h} w_{c'ci'j'} x_{c,i+i',j+j'},$$

for input x and weights $w \in R^{C' \times C \times k_w \times k_h}$. To process video data across time spatiotemporal convolutional neural networks (STCNN) are in practice. Therefore, expression for STCNN is given below:

$$[stcnn(x,w)]_{c'tij} = \sum_{c=1}^{C} \sum_{t'=1}^{k_t} \sum_{i'=1}^{k_w} \sum_{j'=1}^{k_h} w_{c'ci'j'} x_{c,i+i',j+j'}$$

Recurrent neural networks (RNN) improve propagating and learning of information over time steps. We used a type of RNN known as bidirectional GRU (Bi-GRU) [8]. Output of STCNN is fed into Bi-GRU, denoted by $z$. $h_t$ is hidden layer of Bi-GRU. One GRU maps $[z_1, z_2, \ldots z_T] \mapsto [h_1, h_2, \ldots, h_T]$ while second GRU maps $[z_T, \ldots z_2, z_1] \mapsto [h_1, \ldots h_T]$ then $h_t := [\vec{h_t}, \overleftarrow{h_t}]$. Input to GRU is $T$ sequences.

We use connectionist temporal classification (CTC) loss to overcome the problem of training data alignment with target outputs, as described in [3].

### 4. Dataset

To evaluate our model for lip-reading of Urdu speech words and phrases, we constructed a video-speech corpus. Corpus has video-audio recordings 10 participants including both male and female. Corpus contains ten words and ten phrases of Urdu language. Corpus has following words and phrases as shown in the TABLE 1. Detailed information about our dataset is available at [11]. Each participant is requested to repeat each word and phrase ten times. Corpus contains 1000 videos of words and 1000 phrases videos.

**Table 1: Words and Phrases of Urdu Language based Audio-Video Corpus**

| Words | Phrases |
|---|---|
| شروع | تلاش ختم کریں |
| منتخب | معاف کیجڈ گا |
| رابط | میں معافی چاتا وں |
| سمت شناسی | آپکا شکریہ |
| آگڈ بھڑیں | خدا حافظ |
| پیچھڈ جائیں | مجھڈ یڈ کھیل پسند  |
| آغاز | آپ سڈ مل کر خوشی وئی |
| اسلام علیکم | آپکا خیر مقدم  |
| ویب | اپنا خیال رکھئڈ گا |

The recorded videos are cropped to standard size of 100x50, we applied the viola jones [] face detector on each video to detect the face, then we applied the mouth detector to detect the lips and cropped each video so that it only contains the lips movements. This process is illustrated in figure 3.

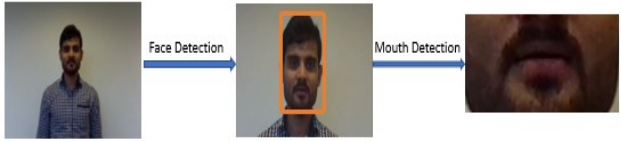

Figure 3: Pre-Processing of videos.

## 5. Implementation Details

For video based lip reading task we have re-implemented the network proposed by [2], the details of the network are given figure. Mainly, the network consists of spatio-temporal convolutional layers (STCNN), bi-directional gated recurrent units (Bi-GRU), and connectionist temporal classification (CTC) loss [3]. The first STCNN takes input of 75 frames at once, and process them together; each STCNN is followed by a max-pooling layer of 2x2 mask.

| Layer | Size / Stride / Pad | Input size | Dimension order |
|---|---|---|---|
| STCNN | 3 × 5 × 5 / 1, 2, 2 / 1, 2, 2 | 75 × 3 × 50 × 100 | $T \times C \times H \times W$ |
| Pool | 1 × 2 × 2 / 1, 2, 2 | 75 × 32 × 25 × 50 | $T \times C \times H \times W$ |
| STCNN | 3 × 5 × 5 / 1, 2, 2 / 1, 2, 2 | 75 × 32 × 12 × 25 | $T \times C \times H \times W$ |
| Pool | 1 × 2 × 2 / 1, 2, 2 | 75 × 64 × 12 × 25 | $T \times C \times H \times W$ |
| STCNN | 3 × 3 × 3 / 1, 2, 2 / 1, 1, 1 | 75 × 64 × 6 × 12 | $T \times C \times H \times W$ |
| Pool | 1 × 2 × 2 / 1, 2, 2 | 75 × 96 × 6 × 12 | $T \times C \times H \times W$ |
| Bi-GRU | 256 | 75 × (96 × 3 × 6) | $T \times (C \times H \times W)$ |
| Bi-GRU | 256 | 75 × 512 | $T \times F$ |
| Linear | 27 + blank | 75 × 512 | $T \times F$ |
| Softmax | | 75 × 28 | $T \times V$ |

Figure 4: LipNet Architecture hyperparameters [2]

After the third STCNN layer the feature vector of 75x96x6x12 is input the Bi-GRU. GRU [7] is a type of RNN that improves upon earlier RNNs by adding cells and gates for propagating information over more timesteps and learning to control this information flow. It is more similar to Long Short-Term Memory (LSTM). Bi-directional RNN were introduced by [8] to increase the amount of input information available to the network. Standard RNN have restrictions as the future information cannot be reached from the current state. On the contrary, in the Bi-directional RNN the future information can be reached from the current state.

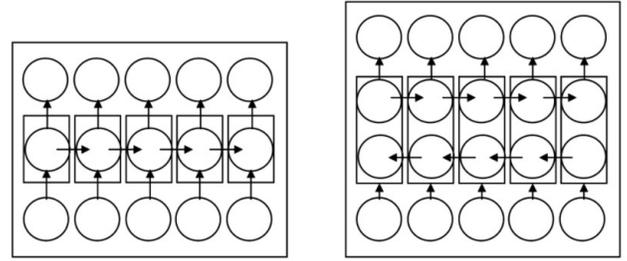

Figure 5: Structure overview (Left Unidirectional RNN, Right Bi-directional RNN) [9]

The speech recognition network only contains the 256 LSTMs cells followed by a fully connected layer and a SoftMax classification layer. The LSTM takes MFCC features as input and the classification layer classify each word as a class. The network architecture is illustrated in figure 2. We implemented both network on tensor flow platform [12].

## 6. Experiments and Results

We have performed several experiments, the first experiment we performed was testing the LipNet [2] on our dataset. We noticed that whatever video we input to LipNet, the output sentence consists of 6 words, this may be because in the GRID dataset, used for training of LipNet, each sentence consists of 6 words. The location of each words is also fixed, for example it is known prior that fourth word would be an English alphabet. Secondly, we also noticed that the output of LipNet varies each time we input the same video. The results of LipNet are shown in Figure 5.

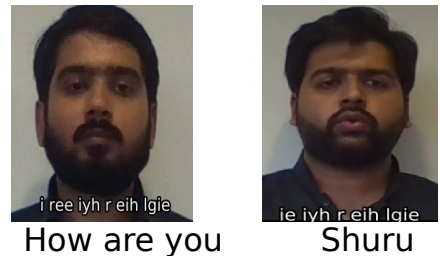

How are you      Shuru

Figure 5: Output of LipNet on different two English sentences and one Urdu word.

After analyzing these results, we coded the same model in tensorflow and tried to train the LipNet on our own dataset, but we could not train the model because of CTC loss. Our model is unable to compute the loss and backpropagate.

The second experiment we performed is training a speech recognition system on only words of our dataset, we pose this problem as classification problem because we only had 10 words in our dataset. We trained two different networks for this task: first one deep neural network and second LSTM based network. Our results showed that LSTM based network performs better than DNN. Moreover, we also trained the both networks on Urdu Digits dataset, taken from CSALT lab. The results of this experiment are tabulated in table 2.

Table 2: Results of Speech Recognition Network

| Dataset | Model | Accuracy |
|---|---|---|
| Words | LSTM Based | **62 %** |
| Words | DNN | 56 % |
| Digits | LSTM Based | **72 %** |
| Digits | DNN | 64 % |

## 7. Conclusion

In this project, we attempted to design an audio-visual lipreading system for Urdu language. We trained two different models for audio and video separately. We successfully trained our audio model for words and digits from Urdu words and digits corpus but we were unable to merge both networks to demonstrate the results of audio-visual lipreading in the noisy environments. Apart from implementation and investigation of models, we contributed a small Urdu language corpus for lipreading. Corpus is consisting of 10 words and 10 phrases, each spoken by 10 users 10 times, in total we recorded and pre-processed the 1000 videos of dataset. In future, our aim is to develop our own model like [1] for lipreading that would be able to take data from two different modalities *i.e.,* speech and video frames as input and outputs a predicted text sequence.